\documentclass[letterpaper]{article} 
\usepackage{aaai23}  
\usepackage{times}  
\usepackage{helvet}  
\usepackage{courier}  
\usepackage[hyphens]{url}  
\usepackage{graphicx} 
\usepackage{natbib}  
\usepackage{caption} 
\usepackage{algorithm}
\usepackage{algorithmic}
\usepackage{microtype}
\usepackage{multirow}
\usepackage{makecell}
\usepackage{amsmath}

\usepackage{newfloat}
\usepackage{listings}
\DeclareCaptionStyle{ruled}{labelfont=normalfont,labelsep=colon,strut=off} 
\floatstyle{ruled}
\newfloat{listing}{tb}{lst}{}
\floatname{listing}{Listing}
\title{On the Calibration and Uncertainty with P\'{o}lya-Gamma Augmentation for \\ Dialog Retrieval Models}
\author{
    Tong Ye\textsuperscript{\rm 1,2},
	Shijing Si\textsuperscript{\rm 1},
	Jianzong Wang\textsuperscript{\rm 1} \thanks{Corresponding author: Jianzong Wang, jzwang@188.com.},
	Ning Cheng\textsuperscript{\rm 1},
	Zhitao Li\textsuperscript{\rm 1},
	Jing Xiao\textsuperscript{\rm 1}
}
\affiliations{\textsuperscript{\rm 1} Ping An Technology (Shenzhen) Co., Ltd.\\
	\textsuperscript{\rm 2} University of Science and Technology of China   
}

\usepackage{bibentry} 

\begin{document}

\maketitle

\begin{abstract}
Deep neural retrieval models have amply demonstrated their power but estimating the reliability of their predictions remains challenging. Most dialog response retrieval models output a single score for a response on how relevant it is to a given question. However, the bad calibration of deep neural network results in various uncertainty for the single score such that the unreliable predictions always misinform user decisions. To investigate these issues, we present an efficient calibration and uncertainty estimation framework PG-DRR for dialog response retrieval models which adds a Gaussian Process layer to a deterministic deep neural network and recovers conjugacy for tractable posterior inference by P\'{o}lya-Gamma augmentation. Finally, PG-DRR achieves the lowest empirical calibration error (ECE) in the in-domain datasets and the distributional shift task while keeping $R_{10}@1$ and MAP performance. 
\end{abstract}

\section{Introduction}
Dialog response retrieval models based on deep neural networks have shown impressive results on multiple benchmarks
\cite{gu2020speaker,lu2020improving,whang2021response}. However, the predictions from these models always fail to provide appropriate answers when deploying into real-world applications. For example, popular dialog agents always show users with incorrect predictions if questions fall outside of the training distribution, which could mislead their decisions. Therefore, an ideal model should abstain when they are likely to be error. The simplest solution is to provide the corresponding confidence estimation so that predictions with low confidence can be abstained.
This problem is also defined as \textbf{Model Calibration}: making sure the confidence of the prediction is well correlated with the actual probabilities of correctness.

Generally, retrieval-based dialog models consider their estimation of a response’s relevance as a deterministic score, which can be subject to over confidence issues, i.e. are badly calibrated \cite{guo2017calibration}.
Existing works usually quantify this uncertainty over predictions through a distribution of possible scores to achieve calibration \cite{cohen2021not}. Specifically, the mean of the distribution represents the model’s prediction while its corresponding variance captures the model’s uncertainty. Consequently,  a high variance could imply that the model is not sure of the prediction and should abstain, even if it is rated among the top hits.

Two of the principled methods to calculate the predicted uncertainty for a dialog model are Deep Ensemble and Bayesian methods. Bayesian methods \cite{cohen2021not} place a prior distribution over model parameters and Deep Ensemble \cite{penha2021calibration} usually trains independently multiple models. They are challenging to implement at the industrial scale due to their high inference cost and huge memory requirements. This inspires us to investigate principled approaches that only need a determinate deep neural network for high quality uncertainty estimation.

Gaussian processes (GPs) \cite{rasmussen2003gaussian} are flexible models that perform well in varieties of tasks. Different from existing works, GPs belong to non-parametric Bayesian approaches, which only need to learn a few hyperparameters. When combined with Gaussian likelihoods, GPs obtain closed form expressions for the predictive and posterior distributions \cite{DBLP:conf/iclr/SnellZ21}, which alleviate the computational defects of Bayesian methods with cubic scaled examples. Moreover, GPs are easily combined with a single deep neural network without training independently multiple models. However, the GPs are challenging to scale to large datasets for classification, this is partially due to the fact that the target variable's categorical distribution leads to a non-Gaussian posterior and we can not obtain the marginal likelihood in a closed form. An especially intriguing family of methods is adding additional P\'{o}lya-Gamma variables to the GPs model \cite{polson2013bayesian} to recover it when the original model is marginalized out.

In this study, we are committed to investigating a simple and efficient approach PG-DRR by combining P\'{o}lya-Gamma Augmentation for calibrating Dialog Response Retrieval models. Specifically, we add a neural Gaussian process layer to a deterministic deep neural network to achieve better calibration. Importantly, we use the P\'{o}lya-Gamma (PG) augmentation to recover conjugacy for tractable posterior inference and use Gibbs sampling to collect samples from the posterior in order to improve the parameters of the mean and covariance functions. Besides, we theoretically verified why PG-DRR can be calibrated.
The significant contributions of this paper are as follows:

\begin{itemize}
	\item We propose an efficient framework PG-DRR for a deterministic dialog response ranking model to estimate uncertainty. And we yield the lowest ECE in two in-domain datasets and the distributional shift task while keeping $R_{10}@1$ and MAP performance.
	\item We innovatively estimate uncertainty in dialog retrieval tasks with a Gaussian Process layer with the P\'{o}lya-Gamma augmentation. In addition, we theoretically analyze that PG-DRR can achieve calibration. 
	\item We conduct extensive experiments to verify that PG-DRR significantly calibrates well while maintaining performance. Besides, ablation study analyzes the relative contributions of the kernel function and the model architecture of PG-DRR to the effectiveness improvement.
\end{itemize}

\section{Related Works}
\subsection{Calibration in Dialog Retrieval}
Most recent approaches for ranking tasks in dialog system have focused primarily on discriminative methods using neural networks that learn a similarity function to compare questions and candidate answers, which has been shown that DNNs result in high calibration errors. 
Therefore, there has been growing research interest in quantifying predictive uncertainty in deep dialog retrieval networks. \cite{zhu2009risky} first investigates the retrieval uncertainty and considers the probabilistic language model's variance is a risk factor to significantly boost performance. 
However, the majority of recent studies on dialog uncertainty put more emphasis on the dialog management aspect \cite{tegho2017uncertainty,gavsic2013gaussian,roy2000spoken,van2020knowing} and few approaches have directly incorporated a retrieval model's uncertainty. 

Deep Ensemble and Monte Carlo (MC) Dropout have emerged as two of the most prominent and practical uncertainty estimation methods for dialog retrieval networks. 
\cite{feng2020none} covers the value of identifying uncertainty in end-to-end dialog jobs and using Dropout to identify questions that cannot be answered. \cite{penha2021calibration} investigates the effectiveness of MC Dropout and Deep Ensemble for calibration in the conversational response space under BERT. 
\cite{cohen2021not} suggests an effective Bayesian framework using a stochastic process to convey model's confidence.  Unfortunately, as retrieval models expand in complexity and size, MC Dropout and Deep Ensemble become computationally expensive, posing a significant challenge given the prevalence of the pre-trained architectures like BERT\cite{durasov2021masksembles}.
In addition, \cite{pei2022transformer} constructs a stochastic self-attention mechanism to capture uncertainty, which means that it needs retraining the models for the different downstream tasks.
Therefore, it is urgent to explore an efficient but straightforward technique to quantify the uncertainty in the determinate neural retrieval models.

\subsection{P\'{o}lya-Gamma Augmentation}
It is obvious that the classification likelihood is non-Gaussian. Sequentially, the predictive distributions are also not Gaussian anymore and its closed-form solution is not available. There are several methods offered to overcome this limitation. The classic methods mainly consist of least squares classification \cite{DBLP:journals/jmlr/RifkinK03}, Laplace approximation \cite{DBLP:journals/pami/WilliamsB98}, Variational approaches \cite{DBLP:conf/aistats/MatthewsHTG16} and expectation propagation \cite{DBLP:phd/ndltd/Minka01}. For a thorough introduction of GPs,
we refer readers to \cite{rasmussen2003gaussian}. 

The P\'{o}lya-Gamma augmentation approach \cite{polson2013bayesian}, which introduces auxiliary random variables to recover it when the original model is marginalized out, can model the discrete likelihood in Gaussian Processes and has attracted extensive attention.  \cite{DBLP:journals/neco/GirolamiR06} investigates a Gaussian augmentation for an accurate Bayesian examination of multinomial probit regression models.
According to \cite{DBLP:conf/nips/LindermanJA15}, a logistic stick-breaking representation and P\'{o}lya-Gamma augmentation are used to convert a multinomial distribution into a product of binomials.
\cite{wenzel2019efficient} represents a scalable stochastic variational method for the Gaussian process classification based on the P\'{o}lya-Gamma data augmentation. 
\cite{DBLP:conf/uai/Galy-FajouWDO19} introduces a logistic-softmax likelihood for multi-class classification and employs Gamma, Poisson and P\'{o}lya-Gamma augmentation in order to obtain a conditionally conjugate model. 
In addition, \cite{DBLP:conf/iclr/SnellZ21} combines P\'{o}lya-Gamma augmentation with the one-vs-each softmax approximation in a novel way and presents a Gaussian process classifier, which demonstrated well calibration. \cite{DBLP:conf/icml/AchituveNYCF21} proposes a GP-Tree framework for multi-class classification based on P\'{o}lya-Gamma augmentation and allows the posterior inference via the variational inference approach or the Gibbs sampling technique.
Therefore, we wish to laverage the conjugacy of the P\'{o}lya-Gamma augmentation to yield better calibrated and more accurate models in dialog tasks.

\begin{figure*}
	\centering
	\includegraphics[width=0.9\linewidth,height=0.22\textheight]{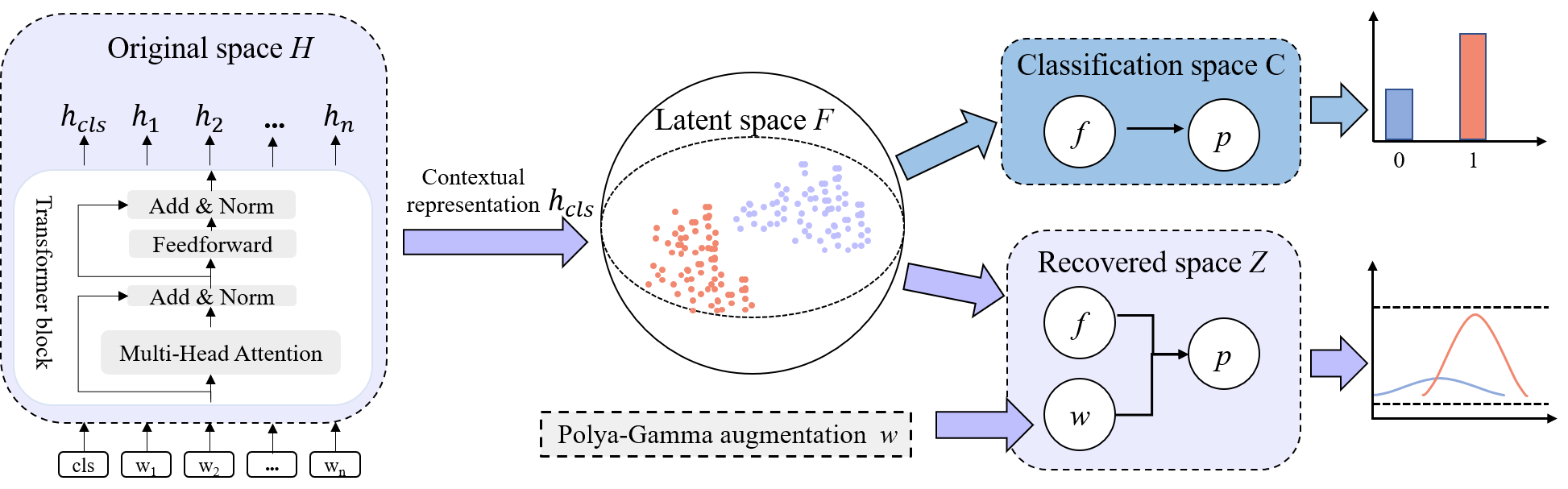}
	\caption{An illustration of PG-DRR prediction models for dialog response retrieval.}
	\label{fig1}
\end{figure*}

\section{Methodology}
In this section, we first recall key properties of P\'{o}lya-Gamma distributions before deriving the augmented model likelihood. We next introduce our method PG-DRR for PG-based Gaussian process classification in detail. All the framework is visually represent in Fig \ref{fig1}.
\subsection{Preliminaries}
The P\'{o}lya-Gamma augmentation method was developed to overcome the Bayesian inference issue in logistic models. \cite{polson2013bayesian}. The P\'{o}lya-Gamma distribution $w\sim PG(b,c)$ with parameters $b>0$ and $c\in R$ can be expressed as a form of Gamma distributions that are infinitely convolved:
\begin{align}
	w\overset{\underset{\mathrm{D}}{}}{=}\dfrac{1}{2\pi^2} \sum_{k=1}^\infty \dfrac{Gamma(b,1)}{(k-1/2)^2+c^2/(4\pi^2)}, 
\end{align}
where $Gamma(b,1)$ indicates Gamma distributions and $\overset{\underset{\mathrm{D}}{}}{=}$ stands for equality of distribution. 
When $w$ follows the P\'{o}lya-Gamma distribution $w\sim PG(b,0)$, the fundamental integral identity holds for $b>0$:
\begin{align}
	\dfrac{(e^\psi)^a}{(1+e^\psi)^b}=2^{-b}e^{\lambda\psi} E_w[e^{-w\psi^2/2}],\label{eq2}
\end{align}
where $\lambda=a-b/2$. 

For the binary classification problems $y\in \{0,1\}^N$, a vector of logits is $\psi\in R^N$. When $a=y$, $b=1$ and $\lambda=y-1/2$ in the Eq. \ref{eq2}, the likelihood can be written as:
\begin{equation}
	\begin{aligned}
		p(y|\psi)&=\sigma(\psi)^{y}(1-\sigma(\psi))^{1-y}= \dfrac{(e^{\psi})^{y}}{1+e^{\psi}} \\
	&=\frac{1}{2}e^{\lambda\psi}\int_{0}^{\infty} e^{-w\psi^2/2}p(w)\, {\rm d}w,
	\label{eq3}
	\end{aligned}
\end{equation}
where the $\sigma(\cdot)$ denotes the sigmoid function. The Eq. \ref{eq3} will be augmented with $w$ such that marginlizing can be used to recover original likelihood.

\subsection{Contextual Encoder}
Given a training set denoted as a triples $D=\{(U_i$, $A_i,$ $y_i)\}_{i=1}^M$, $U_i=\{u_1, u_2, \ldots, u_t\}$ is a dialog context consisting of $t$ utterances and $y_i$ is the response relevance labels $y_i\in \{0,1\}$. Answer candidates is denoted as $A_i=\{a_1, a_2, \ldots, a_r\}$ including $r$ responses.
The input sentence is prepared as $x_i=\{[CLS]$, $u_1$, $\ldots$, $u_t$, $[SEP],a_r\}$. The special token $[CLS]$ denotes the beginning of the sequence and $[SEP]$ denotes the separate the response from the contexts. 
Following BERT \cite{devlin2018bert}, we split the input into words by the same WordPiece tokenizer \cite{DBLP:journals/corr/WuSCLNMKCGMKSJL16} and embeddings for sub-word was obtained by summing up position embedding, word embedding and segment embedding.
Subsequently, BERT is utilized to learn the contextual representation $f(u_i,a_i)$ based on the $[CLS]$ token in a pointwise manner.

\subsection{A Gaussian Process Layer}
The Gaussian process is made up of random variables and every finite subset would follow a multivariate normal distribution.
We can denote a Gaussian process as $g(x)\sim GP(m(x),K(x,x'))$, where $m(x)=E(g(x))$ and $K(x,x')=Cov[g(x),$ $g(x')]$ respectively represents the mean function and the covariance kernel function \cite{rasmussen2003gaussian}.

Supposing the output layer $g:F\rightarrow Z$, PG-DRR replaces the typical dense output layer with a Gaussian process (GP). Specifically, according to the $M$ training samples $D=\{(x_i, y_i)\}_{i=1}^M$, the samples will pass the contextual encoder and obtain a latent representation $h_i=h_{\theta}(x_i)$, where $\theta$ is the weight of the contextual encoder.
The Gaussian process output layer $g=[g(h_1)$, $g(h_2)$, $\ldots,g(h_N)]$ follows a multivariate normal distribution: $g$: $g\sim MVN(0,K(h,h'))$,  
We set radial basis function (RBF) as the kernel function in this paper. Therefore, the formal definition of a GP layer  is denoted as follows:
\begin{equation}
	\begin{aligned}
		&g \sim MVN(0,K),
		K_{i,j}=\sigma^2 exp(-{\left \| h_i-h_j \right \|_2^2}/{2l^2})\label{eq4}
	\end{aligned}
\end{equation}
where $l$ and $\sigma$ are the length and kernel scale parameters \cite{jankowiak2021scalable}, respectively.  

Therefore, the joint likelihood of the latent GP and the label are as follows:
\begin{align}
	p(y|x,g)=\prod_{i} p(y_i|x_i,g_i)
\end{align} 

For a latent function $g$, the posterior distribution $p(g|x,y)$ may be expressed as follows:
\begin{align}
	p(g|x,y)=\frac{p(g)p(y|x,g)}{p(y|x)}
\end{align}

Since the likelihood probability is non-Gaussian, the posterior $p(g|x,y)$ can not be computed analytically. 
Following \cite{wenzel2019efficient}, we build an approximate gradient estimator upon Fisher's identity. We denote $w$ as the P\'{o}lya-Gamma random variables. The full classification's marginal likelihood term is as follows:
\begin{equation}
	\begin{aligned}
		p(y_i|x_i)=\int p_{\theta}(y_i|x_i,w_i)p(w_i)\, {\rm d}w\label{eq7}
	\end{aligned}
\end{equation}

According to the P\'{o}lya-Gamma augmentation and Eq. \ref{eq3}, $p_{\theta}(y_i|x_i,w_i)$ follows a Gaussian distribution in terms of proportion.
\begin{equation}
	\begin{aligned}
		&p_{\theta}(y_i|x_i,w_i)\propto e^{-w_ig_i^2/2}e^{(y_i-1/2)g_i} \\
		&\propto N(diag(w)^{-1}(y_i-1/2)|g,diag(w)^{-1})
	\end{aligned}
\end{equation}

Consequently, the conditional likelihood $p(y|g,w)$ and the Gaussian prior $p(g)$ are conjugate, which leads to the posterior $p(g|y,w)$ also following a Gaussian distribution.
In order to generate samples from this distribution $(g_{(t)},w_{(t)})$, we can readily apply Gibbs sampling \cite{douc2014nonlinear} and calculate the conditional posteriors $p(g|y,w)$ and $p(w|g)$. According to \cite{wenzel2019efficient}, When the auxiliary latent variables $w$ are taken into account, the posterior over the latent function values $g$ is calculated as follows:
\begin{align}
p(w|y,g)&=PG(1,g),\\
p(g|y,w)&=N(g|(K^{-1}\mu+k)\Sigma,\Sigma),\\
\Sigma&=(K^{-1}+diag(w))^{-1}
\end{align}

\subsection{Training Loss}
The following is a possible way to express the log marginal likelihood:
\begin{equation}
	\begin{aligned}
		L&=logp_{\theta}(y|x)
		=log \int p(w)p_{\theta}(y|x,w)\, {\rm d}w \\
		&=log \int p(w)\int p_{\theta}(g|x)p(g|y,w) {\rm d}g {\rm d}w
	\end{aligned}
\end{equation}
By utilizing posterior samples $w\sim p_{\theta}(w|X,Y)$, it is possible to measure the gradient of the log marginal likelihood. 
The samples of $w$ from Gibbs chains serve as the foundation for the stochastic training target in actual practice.
The following describes the gradient estimator upon Fisher's identity.
\begin{equation}
	\begin{aligned}
		\nabla_{\theta} L&=\int p_{\theta}(w|x,y)\nabla_{\theta}log p_{\theta}(y|x,w) {\rm d}w \\
		&\approx\dfrac{1}{N}\sum_{n=1}^{N}\nabla_{\theta}log p_{\theta}(y|w_{n},x)
	\end{aligned}
\end{equation}
where $w_{1}$, $w_{2}, ...,$$w_{N}$ represent the samples generated by the posterior Gibbs chains.

\noindent
\textbf{Prediction} When predicting, we designate the input and label with $x^*$ and $y^*$, respectively.

\begin{equation}
	\begin{aligned}
		&p(g^*|x,y,w,x^*)=N(g^*|\mu^*,\Sigma^*), and\\
		&\mu^*=(K^*)^T(K+diag(w)^{-1})^{-1}diag(w)^{-1}k,\\
		&\Sigma^*=K^{**}-(K^*)^T(K+diag(w)^{-1})^{-1}K^*
	\end{aligned}
\end{equation}
\begin{equation}
	\begin{aligned}
		p(y^*|x,y,w,x^*)=\int p(g^*|x,y,w,x^*)p(y^*|g^*)\, {\rm d}g^*\label{eq13}
	\end{aligned}
\end{equation}
$K^{**}$ indicates the test point's kernel value and $K^*=K(x,x^*)$. We use 1D Gaussian-Hermite quadrature to calculate the intractable integral in Eq.\ref{eq13}.

\subsection{Theoretical Analysis}
Existing works tend to let the weights Bayesian to capture uncertainty. 
GPs belong to non-parametric Bayesian approaches and only need to learn a few hyperparameters. 
Practical interest lies in the following question: Is PG-DRR still Bayesian enough to correct overconfidence? No surprisingly, the answer is yes.
According to \cite{DBLP:conf/cvpr/0001AB19}, when the test data is far away from the training data, ReLU networks exhibit arbitrarily high confidence. In order to achieve our goal, we just need to show that as the gap between test and training data grows, the model prediction approaches zero.

Following \cite{DBLP:conf/cvpr/0001AB19}, let $g: R^d \rightarrow R$ be a binary Gaussian process classifier defined by $g(h(x))$, where $h:R^n \rightarrow R^d$ is a fixed ReLU network denoted as $h(x)=W^Tx+b$. Let $\sigma(\cdot)$ be the sigmoid function and $N(g|\mu,\Sigma)$ be the Gaussian approximation of the last layer's outputs with eigenvalues of $\Sigma^*$ as $\lambda_1\le...\le\lambda_r$. Then for any input $x^*\in R^n$, \\
\begin{equation}
	\begin{aligned}
		p(y^*|x,y,w,x^*)&=\int p(g^*|x,y,w,x^*)p(y^*|g^*)\, {\rm d}g^*\\
		&\approx \int\Phi(\sqrt{\pi/8}g^*)N(g^*|\mu^*,\Sigma^*) {\rm d}g^*	\\
		&=\Phi(\dfrac{\mu^*}{\sqrt{8/\pi+\Sigma^*}})\approx\sigma(z(x^*))	
	\end{aligned}
\end{equation}
where $\Phi$ is denoted as the standard Gaussian distribution function and $\Phi(\sqrt{\pi/8}g^*)\approx\sigma(g^*)$. Following the theorem of \cite{DBLP:conf/icml/Kristiadi0H20}, when $\delta>0$ and as $\delta\rightarrow\infty$, we found:
\begin{equation}
	\begin{aligned}
		\lim_{\delta \to \infty} \sigma(|z(\delta x^*)|)=\lim_{\delta \to \infty} \Phi(\dfrac{\delta\mu^*}{\sqrt{8/\pi+\delta^2\Sigma^*}})\\
		=\lim_{\delta \to \infty} \sigma (\dfrac{\delta\mu^*}{\sqrt{1+\delta^2\pi/8\Sigma^*}})=\sigma(\dfrac{\parallel\mu^*\parallel}{\sqrt{\pi/8\Sigma^*}}) 
	\end{aligned}
\end{equation}

The value goes to a quantity that only depends on the covariance and mean of the Gaussian process. 
This finding suggests that by manipulating the Gaussian, it is possible to move the confidence closer to the uniform further from the training locations.  
We can conclude that The PG-DRR is Bayesian enough to calibrate models and collect uncertainty information.

\subsection{Learning Algorithm}
Algorithm 1 summarizes our learning algorithm for marginal likelihood.
\begin{algorithm}
	\renewcommand{\algorithmicrequire}{\textbf{Input:}}
	\renewcommand{\algorithmicensure}{\textbf{Output:}}
	\caption{PG-DRR Learning}
	\label{alg1}
	\begin{algorithmic}[1]
		\REQUIRE Training data $D$, the weight of networks $\theta$, learning rate $\eta$, number of steps $T$, number of parallel Gibbs chains $N$ 	
		\STATE Initialization parameters $\theta$ randomly
		\REPEAT
		\STATE batch samples $(X,Y)\sim D$
		\FOR{$n=1$ to $N$}
		\STATE     $w_n^{0}\sim PG(1,0)$, $g_{n}^{0}\sim p_{\theta}(g|X)$
		\FOR{$t=1$ to $T$}
		\STATE $w_n^{t}\sim PG(1,g_{n}^{t-1})$
		\STATE  $g_n^{t}\sim p_{\theta}(g|X,Y,w_n^{t})$
		\ENDFOR
		\ENDFOR  
		\STATE   Update $\theta \leftarrow \theta+\tfrac{\eta}{N}\sum_{n=1}^{N}\nabla_{\theta}log p_{\theta}(Y|X,w_n^{T})$
		\UNTIL convergence
		
	\end{algorithmic}  
\end{algorithm}

\section{Experiment}
In this section, we first introduce the datasets and the evaluation metrics which we use to quantitatively compare PG-DRR against MC Dropout and Deep Ensemble. We then compare our method with both in-domain and cross-domain datasets to verify the calibration, efficiency and robustness of PG-DRR. Finally, we discuss the computational and memory complexity.
\subsection{Dataset}
We utilize two large-scale conversational response ranking datasets for our in-domain experiments: 
\begin{itemize}
	\item \textbf{MS Dialog}\cite{qu2018analyzing}, crawled from over 35K information seeking conversations from the Microsoft Answer community, contains 246,000 context-response pairs. For each utterance, it is accompanied with a list of 10 potential responses, only one of which is accepted as the right response.
	\item \textbf{MANtIS}\cite{penha2019introducing}: based on the Stack Exchange community question-answering website and has 1.3 million context-response pairs with 14 distinct domains. For each context, We create 10 negative sampled instances by replacing the provider's most recent statement with a negative sample drawn from BM25 using the correct response as the query. 
\end{itemize}

Besides using the uncertainty estimation for in-domain datasets, we also train the model using the training set from one dataset, i.e. train set and predict it on the test set of a different dataset, which is also know as domain generalization or distributional shift tasks.
\begin{itemize}
	\item \textbf{MS Dialog$\rightarrow$MANtIS} This distributional shift scenario takes the training set of MS Dialog and to validate and test on the MANtIS.
	\item \textbf{MANtIS$\rightarrow$MS Dialog} This distributional shift scenario takes the training set of MANtIS and to validate and test on the MS Dialog.
	
\end{itemize}

\subsection{Metrics}
\textbf{Recall@1}: For retrieval performance evaluation, we employ the recall at 1 out of 10 candidates named $R_{10}@1$, which consists of 1 ground-truth response and 9 candidates randomly selected from the test set.

\noindent
\textbf{Mean Average Precision (MAP)}: To evaluate average precision across multiple queries, we use the MAP. Simply put, it is the mean of the accuracy average over all queries.

\noindent
\textbf{Empirical Calibration Error (ECE)}: To evaluate the calibration efficiency of dialog retrieval tasks, we utilize the Empirical Calibration Error (ECE)\cite{naeini2015obtaining}. It can assess the relationship between expected probability and accuracy.
We partition the range $[0,1]$ into $C$ bins that are evenly spaced apart. 
By calculating the weighted average of the absolute difference between the accuracy  $A_i$ and confidence $B_i$ of each bin, the ECE may be roughly calculated as: $ECE=\sum_{i=1}^{C} \frac{|B_i|}{N}|A_i-mean(B_i)|$. 
We set $C=10$.

\subsection{Baselines}
We compare several recent methods for qualifying uncertainty in conversation response ranking tasks. 

\noindent
\textbf{BERT}\cite{devlin2018bert}: a pre-trained BERT (bert-base-cased) that fine-tuned on the dialog response retrieval task. 

\noindent
\textbf{MC Dropout}\cite{penha2021calibration}: a BERT-based model employing dropout at both train and test time to approximate Bayesian inference and generating a predictive distribution after 10 forward passes.

\noindent 
\textbf{Deep Ensemble}\cite{penha2021calibration}: integrating predictions of any 5 BERT models with dense output layers. We denote as Ensemble for simplification.

\noindent
\textbf{SNGP}\cite{liu2020simple}: A modified version of BERT that the output layer was replaced by a Gaussian Process layer and weighting normalization step was add to training loop. 

\begin{table*}[]
	\centering
	\caption{Average performance ($R_{10}@1$, MAP) and standard deviation over 5 seeds for dialog response retrieval tasks. The best results are highlighted in bold.}
	\scalebox{1.1}{
	\begin{tabular}{llccccc}
		\Xhline{1.2pt}
		&Metric     & BERT            & MC Dropout      & Ensemble                 & SNGP            & PG-DRR                  \\ \Xhline{1.2pt}
		\multirow{2}{*}{MS Dialog} & $R_{10}@1$  &0.658$\pm$0.010 & 0.644$\pm$0.009 & \textbf{0.678$\pm$0.004} & 0.600$\pm$0.036 & 0.656$\pm$0.046          \\
		& MAP & 0.782$\pm$0.007 & 0.776$\pm$0.005 & \textbf{0.798$\pm$0.003} & 0.744$\pm$0.022 & 0.784$\pm$0.032            \\ \hline
		\multirow{2}{*}{MANtIS}   & $R_{10}@1$ & 0.556$\pm$0.028  & 0.562$\pm$0.026 & 0.553$\pm$0.012          & 0.545$\pm$0.050 & \textbf{0.606$\pm$0.016} \\
		& MAP & 0.683$\pm$0.021 & 0.688$\pm$0.019 & 0.681$\pm$0.008          & 0.680$\pm$0.031 & \textbf{0.731$\pm$0.011} \\ 
		 \Xhline{1.2pt}
	\end{tabular}}
	
	\label{tab:my-table1}
\end{table*}

\begin{figure*}
	\centering
	\includegraphics[width=1\linewidth,height=0.25\textheight]{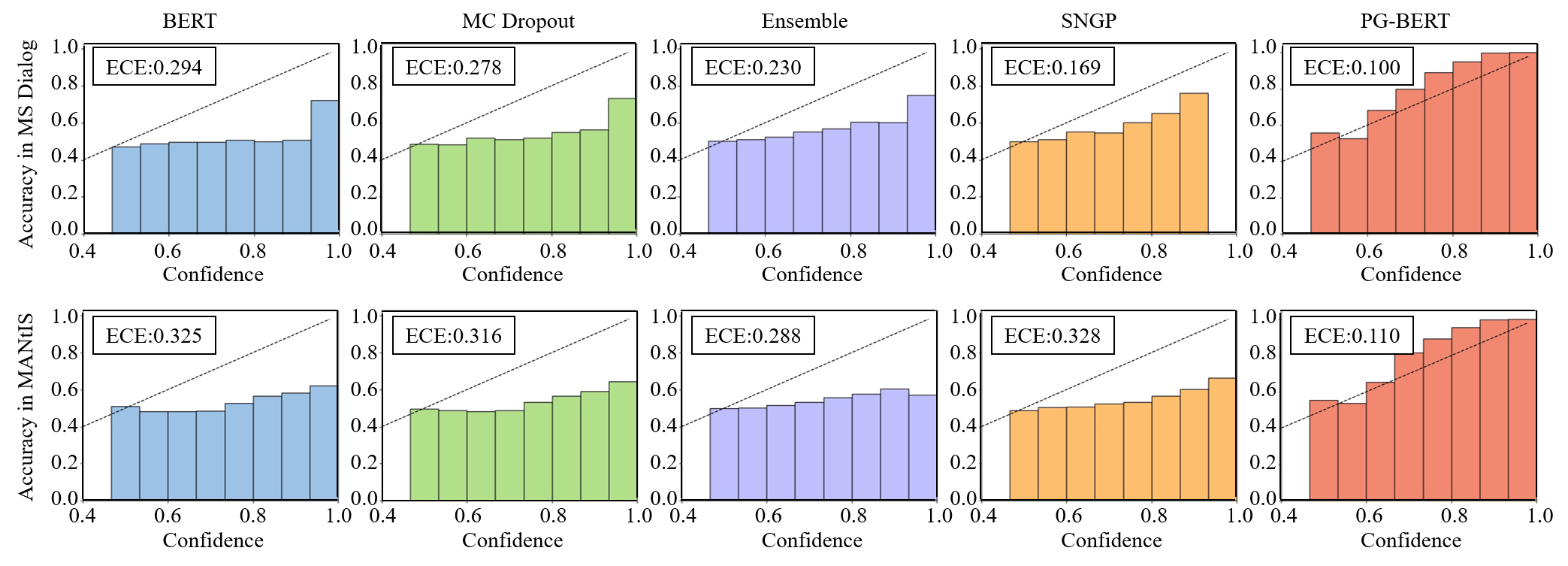}
	\caption{Reliability diagrams and expected calibration error (ECE) on the MS Dialog and MANtIS dataset.}
	
	\label{fig2}
\end{figure*}

\begin{table}[]
	\caption{Parameters and wall-clock time (in millisecond) of inference over test sets.}
	\centering
	
	\scalebox{1}{
		\begin{tabular}{lcc}
			\Xhline{1.2pt}
			& Parameters (M) & Inference time (ms) \\ \hline
			BERT       & 108.31           &11.48                \\
			MC Dropout &108.31            &111.28                \\
			Ensemble   &514.56            &57.39                \\
			SNGP       &118.87            &12.82                \\
			PG-DRR         &108.90            &17.09     \\
			\Xhline{1.2pt}
	\end{tabular}}	
	\label{tab:my-table3}
\end{table}

\begin{table*}[]
	\centering
	\caption{Average calibration (ECE) and performance ($R_{10}@1$, MAP) over 5 seeds for distributional shift tasks. $"\uparrow"$ represents the higher the better while $"\downarrow"$ means the opposite.}
	\scalebox{1.0}{
		\begin{tabular}{lcccccc}
			\Xhline{1.2pt}
			& \multicolumn{3}{c}{MS Dialog$\rightarrow$MANtIS} & \multicolumn{3}{c}{MANtIS$\rightarrow$MS Dialog}   \\ \cline{2-7}
			& $R_{10}@1$$\uparrow$     & MAP$\uparrow$     & ECE$\downarrow$     & $R_{10}@1$$\uparrow$      & MAP$\uparrow$     & ECE$\downarrow$       \\ \Xhline{1.2pt}
			BERT       &0.378$\pm$0.015           &0.537$\pm$0.012                      &0.343$\pm$0.035         &0.424$\pm$0.047           &0.599$\pm$0.035         &0.514$\pm$0.037                            \\
			MC Dropout &0.358$\pm$0.032           &0.523$\pm$0.023                      &0.328$\pm$0.045                  &0.409$\pm$0.023           &0.590$\pm$0.020         &0.497$\pm$0.040                              \\
			Ensemble   &0.405$\pm$0.007           &0.558$\pm$0.005                   &0.331$\pm$0.037         &\textbf{0.457$\pm$0.019}           &\textbf{0.625$\pm$0.013}        &0.503$\pm$0.037                            \\
			SNGP       &0.340$\pm$0.042           &0.514$\pm$0.032                   &0.307$\pm$0.011         &0.333$\pm$0.126           &0.520$\pm$0.107         &0.485$\pm$0.017                               \\
			PG-DRR    &\textbf{0.530$\pm$0.049}           &\textbf{0.673$\pm$0.038}         &\textbf{0.101$\pm$0.003}         &0.330$\pm$0.014           &0.553$\pm$0.009         &\textbf{0.091$\pm$0.002}           \\ \Xhline{1.2pt}                  
	\end{tabular}}
	
	\label{tab:my-table2}
\end{table*}

\subsection{Implementation Details}
We employ $BERT$, which has a hidden state dimension of 768 and consists of 12 Transformer blocks with 12 attention heads, as the backbone.
We utilize the Adam optimizer and set dropout probability to 0.1. Following recent research \cite{gu2020speaker} that employed finetuned BERT for dialog response ranking, we choose a response at random from the whole collection of responses as the negative samples when training. 

In our experiments, we set the learning rates to 1e-5 for baseline methods, and personal learning rates to 3e-3 for GP layer (GP layer is not fine-tuned and therefore need bigger rates). RBF kernel function is configured to use fix lenght scale of 1 and 8 output scale. During training and evaluation, we both use 10 steps of Gibbs and 30 parallel Gibbs chains to reduce variance. We use the same set of hyper parameters across different model architecture for deterministic variant. In the MS Dialog dataset, we train our model with a batchsize of 16 for 3 epochs while 1 epoch for the MANtIS dataset. We used Pytorch and models were trained on 1 Tesla V100 with 16G memory \footnote{Detailed experimental codes can be found at https://github.com/pingantechnlp/PG-DRR.}.

\section{Results}
In this section, we present our proposed PG-DRR in terms of both performance and calibration. Our aim is to compare the calibration among a variety of methods. Therefore, we need to develop new benchmark evaluations: retrieval performance, the calibration, and the robustness. In order to achieve our goal, we could not just take the accuracy findings from previous studies, but instead, each of these baselines requires being trained from scratch.

\subsection{Performance, Calibration and Efficiency}
\textbf{Performance} We first report the performance of PG-DRR in comparison with other baselines on the MS dialog and MANtIS datasets by evaluating the metrics of $R_{10}@1$ and MAP. The results are represented in the Table \ref{tab:my-table1}.

As shown, PG-DRR is competitive with that of a deterministic network BERT, and outperforms the other single-model calibration approaches like MC Dropout and SNGP on the MS Dialog dataset. In addition, PG-DRR is competitive with the Ensemble, which is reduced by less than 2 \%. For MANtIS,  PG-DRR yields strong retrieval results, outperforming all the deterministic and calibration approaches. This indicates that PG-DRR generally achieves competitive prediction performance under in-domain conditions.

\noindent
\textbf{Calibration} We turn to the calibration for dialog retrieval tasks, which is crucial for a dialog system to delay to make a wise choice when there is insufficient information. We chose the most commonly used metric ECE for evaluating calibration. The results can be found in the Figure \ref{fig2}.

Obviously, The confidence of BERT, MC Dropout, Ensemble and SNGP is much higher than their accuracy, which means that they are deeply overconfident. Specifically, BERT, which is a vanilla model, usually performs well but is not well calibrated. The calibration of MC Dropout and Deep Ensemble outperform BERT, which verifies that Bayesian models will have better confidence expressiveness, but unfortunately still obtain poor calibration. Compared to the prior methods, the ECE of PG-DRR is almost lowest, which is reduced by almost 6\% and 16\% respectively in two datasets, while the $R_{10}@1$ and MAP are better or less than 1\% decrease. Namely, PG-DRR includes uncertainty information while keeping $R_{10}@1$ and MAP performance in in-domain datasets.

\noindent
\textbf{Efficiency}
Computational cost is One of the most significant obstacles when employing Bayesian to capture uncertainty. We analyze the efficiency of PG-DRR on the number of parameters and the inference time in the MS Dialog dataset in Table \ref{tab:my-table3}. Compared to MC Dropout and Ensemble, PG-DRR has a significant decrease (at least 8 times) on inference time. While not completely free, PG-DRR only adds negligible computational cost to BERT, but it greatly improves the calibration, which facilitates adaptation to other models. In addition, non-parameters Gaussian process significantly reduces the number of parameters compared with Ensemble. 

We believe that GPs belong to non-parametric Bayesian approaches and only need to learn a few hyperparameters, which enables PG-DRR improve uncertainty calibration for dialog response retrieval models. This happens to be coincident with \cite{DBLP:conf/icml/Kristiadi0H20} that a little Bayesian is enough to capture uncertainty information and correct the overconfidence even if it is only placed at the last layer of model when handling binary classification.
\begin{figure}
	\centering
	\includegraphics[width=1\linewidth,height=0.22\textheight]{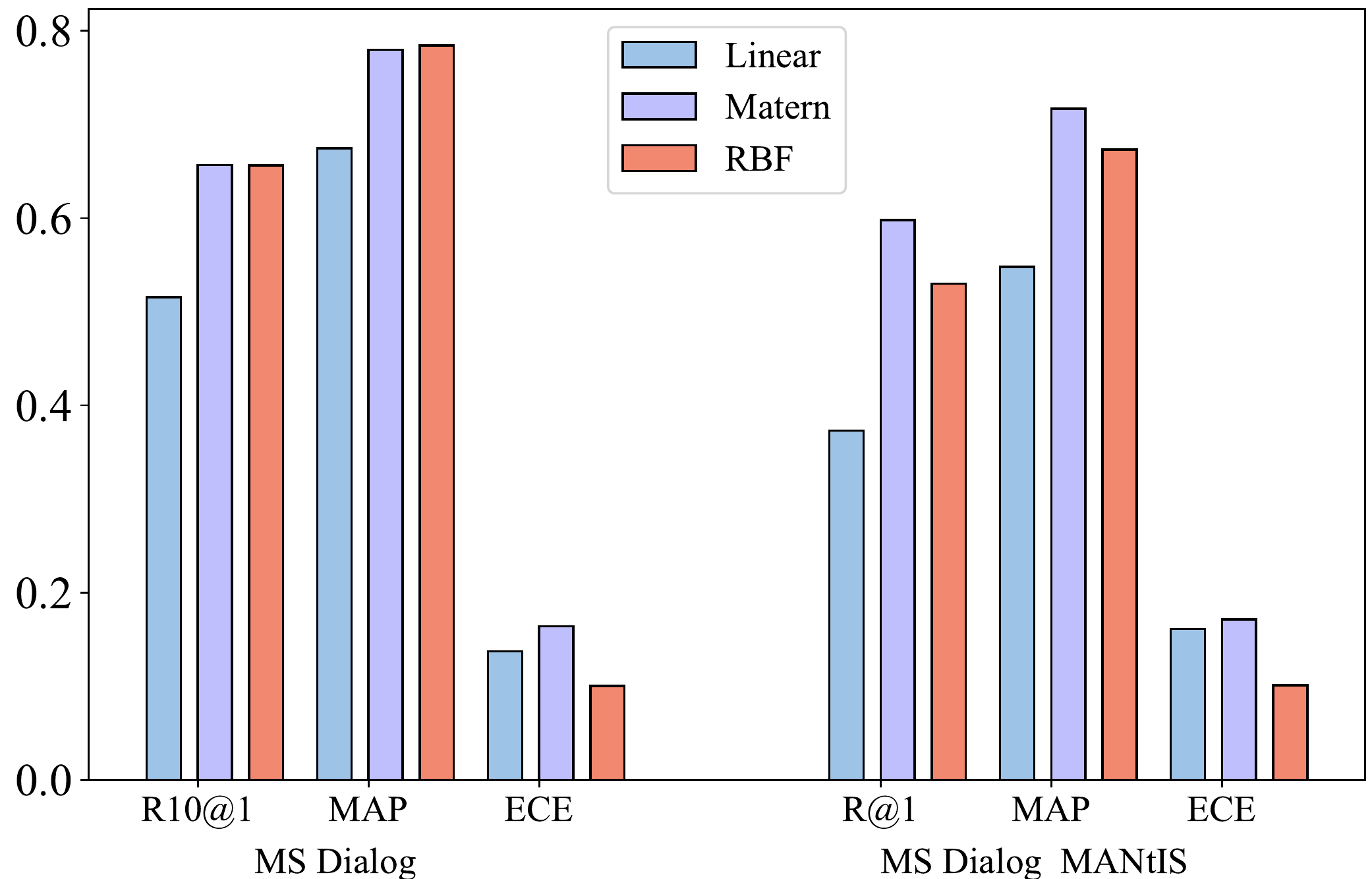}
	\caption{Performance and calibration for response retrieval tasks with different kernel functions.}
	\label{fig3}
\end{figure}

\subsection{Robustness and Generalization}
To evaluate the model’s robustness and generalization, we consider the distributional shift task.

As for the distributional shift task, we develop the model using the test set from a separate dataset after training it on the training set from the first dataset, which is also known as domain generalization tasks. We record all the retrieval performance and calibration in Table \ref{tab:my-table2}. As shown, we observe that SNGP has a lower ECE value than BERT, MC Dropout, and Ensemble, indicating better calibration. Additionally, SNGP achieves competitive performance on the $R_{10}@1$ and MAP metrics. 
Under the SNGP architecture, the PG-DRR achieves a significant drop of at least 20\% to the upper calibration bound, and its $R 10@1$ and MAP are significantly higher or competitive in distribution shift tasks.
This demonstrates that GP-based retrieval models will convey their confidence in distributional shift tasks with more robust expressiveness. According to the results of SNGP and PG-DRR, we find that performance, calibration, and resilience are strongly influenced by the posterior.

\subsection{Ablation Study}
\textbf{Kernel function} The performance of Gaussian process is associated with kernel function, so we frequently discuss how choosing a different kernel affects dialog retrieval for our PG-DRR. According to \cite{DBLP:conf/icml/AchituveNYCF21}, we compare the Linear, Matern and RBF kernel. The results can be found in the Figure \ref{fig3}. It is clear that the RBF kernel performs better in our scenario. We hypothesis that the representation in the embedding space is more mixed. The RBF kernel tend to generate non-linear decision boundaries and thus achieves stronger and better performance.

\noindent
\textbf{Model architecture} We next attempt to disentangle how the differences of the model architecture affect their calibration properties. In this study, we choose the RoBERTa, another famous models in nature language processing, to verify the generalization performance of the PG-DRR method in the in-domain and cross-domain scenario. As shown in Table \ref{tab:my-table4}, We discovered that the RoBERTa-based approaches outperform BERT-based methods in terms of retrieval performance and calibration error. We can assume that model architecture and pretraining may affect model calibration and performance deeply. Regardless of model architectures, PG-DRR can always remain state-of-the-art calibration error, which indicates that PG-DRR can be applied to various model architectures.

\begin{table}[]
	\caption{Effectiveness ($R_{10}@1$, MAP) for dialog response retrieval. $"\uparrow"$ represents the higher the better while $"\downarrow"$ means the opposite.}
	\scalebox{0.88}{
	\begin{tabular}{lccc}
		\Xhline{1.2pt}
		 &R10@1$\uparrow$ & MAP$\uparrow$ & ECE$\downarrow$ \\ \hline
		RoBERTa    &0.632$\pm$0.005       &0.767$\pm$0.003     &0.208$\pm$0.029     \\
		MC Dropout &0.615$\pm$0.021       &0.756$\pm$0.014     &0.198$\pm$0.059     \\
		Ensemble   &0.666$\pm$0.006       &0.790$\pm$0.003     &0.165$\pm$0.018     \\
		SNGP       &0.488$\pm$0.072       &0.669$\pm$0.053     &0.194$\pm$0.021     \\
		PG-DRR         &0.626$\pm$0.024       &0.764$\pm$0.013     &0.110$\pm$0.006    \\ \hline \hline
		&R10@1$\uparrow$ & MAP$\uparrow$ & ECE$\downarrow$ \\ \hline
		RoBERTa    &0.485$\pm$0.025       &0.627$\pm$0.019     &0.385$\pm$0.024     \\
		MC Dropout &0.441$\pm$0.021       &0.594$\pm$0.015     &0.440$\pm$0.024     \\
		Ensemble   &0.532$\pm$0.007       &0.662$\pm$0.005     &0.330$\pm$0.019     \\
		SNGP       &0.384$\pm$0.072       &0.554$\pm$0.060     &0.366$\pm$0.052     \\
		PG-DRR         &0.484$\pm$0.057       &0.632$\pm$0.044     &0.166$\pm$0.018    \\ \Xhline{1.2pt}
	\end{tabular}}
	\label{tab:my-table4}
\end{table}

\section{Conclusion}
In this paper, we present an efficient uncertainty estimation architecture PG-DRR for reliable dialog response retrieval tasks. PG-DRR only adds a neural GP layer to a deterministic DNN to improve the ability of calibration and use the P\'{o}lya-Gamma augmentation to recover conjugacy for tractable posterior inference and optimize the parameters. In addition, Gibbs sampling is used to collect samples from the posterior. We conduct extensive experiment to verify the effectiveness between parameter and inference time. Besides, we explore the relative contributions of the kernel function and the model architecture of PG-DRR to the effectiveness improvement in the ablation study. We believe that Gaussian process modeling is an effective technique for managing calibration and hope that our work will shed more light on efficient Bayesian inference in dialog retrieval scenario.

\section{Acknowledgment}
This paper is supported by the Key Research and Development Program of Guangdong Province under grant No. 2021B0101400003. Corresponding author is Jianzong Wang from Ping An Technology (Shenzhen) Co., Ltd (jzwang@188.com).

\bibliography{aaai23}
\bibliographystyle{aaai23}

\end{document}